\pdfoutput=1
\documentclass[11pt]{article}

\usepackage{acl}

\usepackage{times}
\usepackage{latexsym}
\usepackage{xspace}
\usepackage{amsmath}
\usepackage{tabularx}

\usepackage[T1]{fontenc}
\usepackage[utf8]{inputenc}

\usepackage{microtype}

\usepackage{booktabs}
\usepackage{url}

\usepackage{graphicx}
\usepackage{caption}
\usepackage{subcaption}
\usepackage{tikz}
\usepackage{pgfplots}
\usepgfplotslibrary{fillbetween}
\usetikzlibrary{patterns}
\pgfplotsset{compat=1.8}

\newenvironment{customlegend}[1][]{%
    \begingroup
    \csname pgfplots@init@cleared@structures\endcsname
    \pgfplotsset{#1}%
}{%
    \csname pgfplots@createlegend\endcsname
    \endgroup
}%

\def\addlegendimage{\csname pgfplots@addlegendimage\endcsname}

\newcommand{\modelname}{FiDO\xspace}
\newcommand{\fidlight}{FiD-Light\xspace}

\definecolor{blue}{rgb}{0, 0.46484375, 0.73046875} % blue #0077BB
\definecolor{orange}{rgb}{0.9296875, 0.46484375, 0.19921875} % orange #EE7733
\definecolor{cyan}{rgb}{0.2980, 0.7529, 0.7490} % cyan #33BBEE
\definecolor{red}{rgb}{0.796875, 0.19921875, 0.06640625} % red #CC3311
\definecolor{teal}{rgb}{0, 0.59765625, 0.53125} % teal #009988
\definecolor{magenta}{rgb}{0.9296875, 0.19921875, 0.46484375} % magenta #EE3377
\definecolor{grey}{rgb}{0.73046875,0.73046875,0.73046875} % grey #BBBBBB

\definecolor{fidocolor}{rgb}{0, 0.46484375, 0.73046875} % blue #0077BB
\definecolor{fidcolor}{rgb}{0.9296875, 0.46484375, 0.19921875} % orange #EE7733

\definecolor{crosscolor}{rgb}{0.2980, 0.7529, 0.7490} % cyan

\definecolor{darkgrey}{rgb}{0.33203125,0.33203125,0.33203125} % grey #555555

\definecolor{decodercolor}{rgb}{0.9296875, 0.3984375, 0.46484375} % red #EE6677
\definecolor{encodercolor}{rgb}{0.265625, 0.46484375, 0.6640625} % blue #4477AA

\title{FiDO: Fusion-in-Decoder optimized for stronger performance and faster inference}

\author{Michiel de Jong\thanks{Correspondence to msdejong@usc.edu. Work done at Google Research.} \ \footnotemark[2],~ Yury Zemlyanskiy\footnotemark[3],~ Joshua Ainslie\footnotemark[3],~ Nicholas FitzGerald\footnotemark[3] \\ {\bf Sumit Sanghai\footnotemark[3],}~ {\bf Fei Sha\footnotemark[3],}~ {\bf William W. Cohen\footnotemark[3]}\\
\footnotemark[2] \footnotetext[2]{ } University of Southern California, \footnotemark[3] \footnotetext[3]{ }  Google Research}

\begin{document}
\maketitle
\begin{abstract}
Fusion-in-Decoder (FiD) is a powerful retrieval-augmented language model that sets the state-of-the-art on many knowledge-intensive NLP tasks. However, the architecture used for FiD was chosen by making minimal modifications to a standard T5 model, which our analysis shows to be highly suboptimal for a retrieval-augmented model.  In particular, FiD allocates the bulk of FLOPs to the encoder, while the majority of inference time results from memory bandwidth constraints in the decoder. We propose two simple changes to the FiD architecture to alleviate memory bandwidth constraints, and speed up inference by 
7x. This allows us to use a much larger decoder at modest cost. %Finally, we rebalance compute by employing a much larger decoder at modest cost. 
We denote FiD with the above modifications as \modelname, and show that it strongly improves performance over existing FiD models for a wide range of inference budgets. For example, \modelname-Large-XXL performs faster inference than FiD-Base and achieves better performance than FiD-Large.
\end{abstract}

% method framing, related work, conclusion

\section{Introduction}

A large body of work has demonstrated that language model performance on downstream tasks can be improved by augmenting the model with relevant retrieved text \citep{realm, rag, fid, atlas}. In particular, the Fusion-in-Decoder (FiD) architecture \citep{fid} stands out for strong performance, even outperforming much larger models on many knowledge-intensive tasks \citep{atlas}. However, FiD uses a standard T5 encoder-decoder architecture \cite{t5} which was not designed for use as a retrieval-augmented model. In this work we propose \modelname, a modified FiD architecture optimized for the retrieval-augmented setting.

The FiD decoder is responsible for a difficult task, assimilating information from many passages and reasoning over the information to generate an output. However, because the encoder and decoder are similar size and the encoder is applied to a large number of retrieved passages, FiD devotes an order of magnitude more Floating Point Operations (FLOPs) to the encoder than the decoder. In spite of this, the majority of inference time is actually spent in the decoder, as has been observed in prior work  \citep{fidlight}. This surprising result is shown in Figure~\ref{fig:flops_train_inference}.  Our analysis finds that for typical inference settings the FiD decoder is memory-bandwidth bound \citep{roofline} due to using multi-head cross-attention \citep{attention} over a large input sequence.

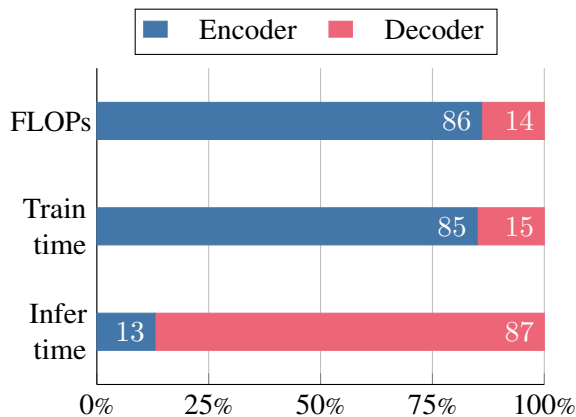
\begin{figure}  
\centering
\begin{tikzpicture}[scale=1.0]
\begin{axis}[
    xbar stacked,
    % bar shift=0pt,
    bar width=14pt,
    enlarge y limits=0.25,
    width=0.97\columnwidth,
    height=0.75\columnwidth,
    major y tick style = transparent,    
    xmajorgrids = true,
    symbolic y coords={Infer time, Train time, FLOPs},
    ytick = data,
    xtick = {0, 25, 50, 75, 100},
    xticklabels = {0\small{\%}, 25\small{\%}, 50\small{\%}, 75\small{\%}, 100\small{\%}},
    yticklabel style={align=right, text width=1cm},
    xmin=0,
    xmax=100,
    axis y line*=none,
    axis x line*=bottom,
    nodes near coords,
    nodes near coords align={anchor=east},
    point meta=rawx,
    nodes near coords style={
        text=white,
    },
    legend columns=2,
    legend cell align=left,
    legend style={
        anchor=south,
        at={(0.49, 1)},
        column sep=2ex,
        yshift=5
    },
]
    \addplot[style={encodercolor,fill=encodercolor,mark=none,opacity=1.0}]
        coordinates {(86,FLOPs) (85,Train time) (13,Infer time)};    
    \addplot[style={decodercolor,fill=decodercolor,mark=none,opacity=1.0}]
        coordinates {(14,FLOPs) (15,Train time) (87,Infer time)};        
    \legend{Encoder,Decoder}        
\end{axis}
\end{tikzpicture}            
\caption{Shows the percentage of FLOPs in forward pass, training time and inference time for the encoder and decoder for a Fusion-in-Decoder model with 40 retrieved passages and batch size 24. The vast majority of FLOPs and training time originate from the encoder, but the decoder is much more expensive for inference.}
\label{fig:flops_train_inference}
\end{figure}
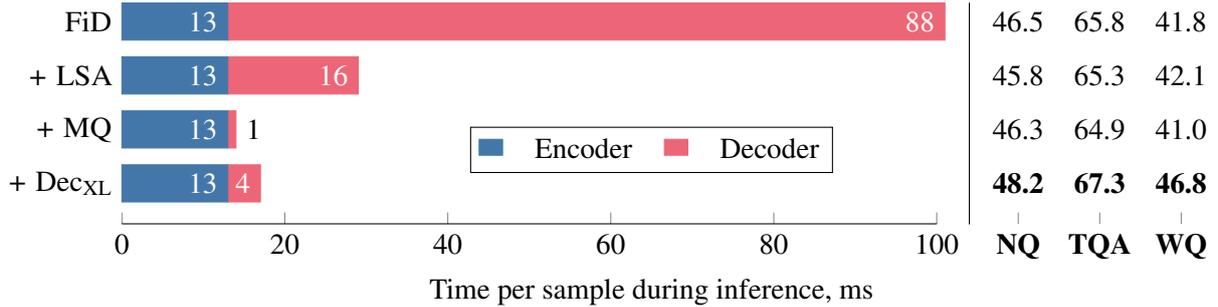
\begin{figure*}  
\centering
\begin{tikzpicture}
\begin{axis}[
    xbar stacked,
    % bar shift=0pt,
    bar width=14pt,
    enlarge y limits=0.25,
    width=0.97\textwidth,
    height=0.3\textwidth,
    major y tick style = transparent,    
    % xmajorgrids = true,
    % xmajorgrids = {0, 10},
    xlabel = {Time per sample during inference, ms},
    symbolic y coords={XLD, MQ, LSA, FiD},
    ytick = data,
    yticklabels = {FiD, + LSA, + MQ, + $\text{Dec}_{\text{XL}}$},
    yticklabel style={align=right, text width=1.3cm},
    xtick = {0, 20, 40, 60, 80, 100, 110, 120, 130},
    xticklabels = {0, 20, 40, 60, 80, 100, \textbf{NQ}, \textbf{TQA}, \textbf{WQ}},
    axis line style={shorten >= 91.5pt},
    xmin=0,
    xmax=130,
    clip=false,
    axis y line*=none,
    axis x line*=bottom,
    legend columns=2,
    legend cell align=left,
    legend style={
        anchor=center,
        at={(0.5, 0.25)},
        column sep=2ex,
        yshift=5
    },
]

    \node [white, left] at (axis cs: 13,FiD) {13};
    \node [white, left] at (axis cs: 13,LSA) {13};
    \node [white, left] at (axis cs: 13,MQ) {13};
    \node [white, left] at (axis cs: 13,XLD) {13};

    \node [white, left] at (axis cs: 101,FiD) {88};
    \node [white, left] at (axis cs: 29,LSA) {16};
    \node [black, right] at (axis cs: 14,MQ) {1};
    \node [white, left] at (axis cs: 17,XLD) {4};
    
    \addplot[style={encodercolor,fill=encodercolor,mark=none,opacity=1.0,text=white}] coordinates {(13,FiD) (13,LSA) (13,MQ) (13,XLD)};    
    
    \addplot[style={decodercolor,fill=decodercolor,mark=none,opacity=1.0,text=black}] coordinates {(88,FiD) (16,LSA) (1,MQ) (4,XLD)};    
    
    % \node [black, yshift=-3] at (axis cs: 56,[normalized]-1) {\textbf{NQ EM}};
    \node [black] at (axis cs: 110,FiD) {46.5};
    \node [black] at (axis cs: 110,LSA) {45.8};
    \node [black] at (axis cs: 110,MQ) {46.3};
    \node [black] at (axis cs: 110,XLD) {\textbf{48.2}};
    
    \node [black] at (axis cs: 120,FiD) {65.8};
    \node [black] at (axis cs: 120,LSA) {65.3};
    \node [black] at (axis cs: 120,MQ) {64.9};
    \node [black] at (axis cs: 120,XLD) {\textbf{67.3}};
    
    \node [black] at (axis cs: 130,FiD) {41.8};
    \node [black] at (axis cs: 130,LSA) {42.1};
    \node [black] at (axis cs: 130,MQ) {41.0};
    \node [black] at (axis cs: 130,XLD) {\textbf{46.8}};
    
    \draw (104,-76) -- (104,335);
    
  \legend{Encoder,Decoder}        
\end{axis}
\end{tikzpicture}            
\caption{\textbf{MAIN RESULT. Layer-sparse cross-attention (LSA) and multi-query (MQ) attention eliminate the bulk of decoder inference cost with minor performance penalty, and the decoder can then be massively scaled up ($\text{Dec}_{\text{XL}}$) with only a modest increase in inference time.} To the left, encoder and decoder inference time per sample on a single TPUv4 with batch size 24 and 40 retrieved passages for variants of base-sized FiD model. To the right, corresponding exact match performance on Natural Questions (NQ), TriviaQA (TQA) and WebQuestions (WQ) dev sets.}
\label{fig:encoder_decoder_inference}
\end{figure*}

Based on this analysis, we propose two sets of architectural changes. We first propose to reduce the cost of cross-attention over retrieved passages by removing most cross-attention layers from the decoder. This reduces cost and yields much smaller losses in performance than \fidlight\citep{fidlight}, the best previously-proposed approach for optimizing FiD. We also replace multi-head attention with multi-query attention \citep{multiquery}. With these modifications the memory-bandwidth bottleneck is eliminated: decoder inference is now orders of magnitude faster and most inference time is spent in the encoder, consistent with the balance of FLOPs between components. 

Finally, we propose to partially rebalance compute towards the decoder by massively scaling decoder size, using a smaller encoder to extract information from retrieved passages and a larger decoder to assimilate the information and reason about the desired output. We refer to the resulting series of models as \modelname (Fusion in Decoder Optimized) and show that \modelname strongly outperforms standard FiD models on the question-answering datasets Natural Questions \citep{nq}, TriviaQA \citep{triviaqa} and WebQuestions \citep{webquestions} for a wide range of inference budgets and settings.  Figure~\ref{fig:encoder_decoder_inference} summarizes some of these results.

\section{Analysis}
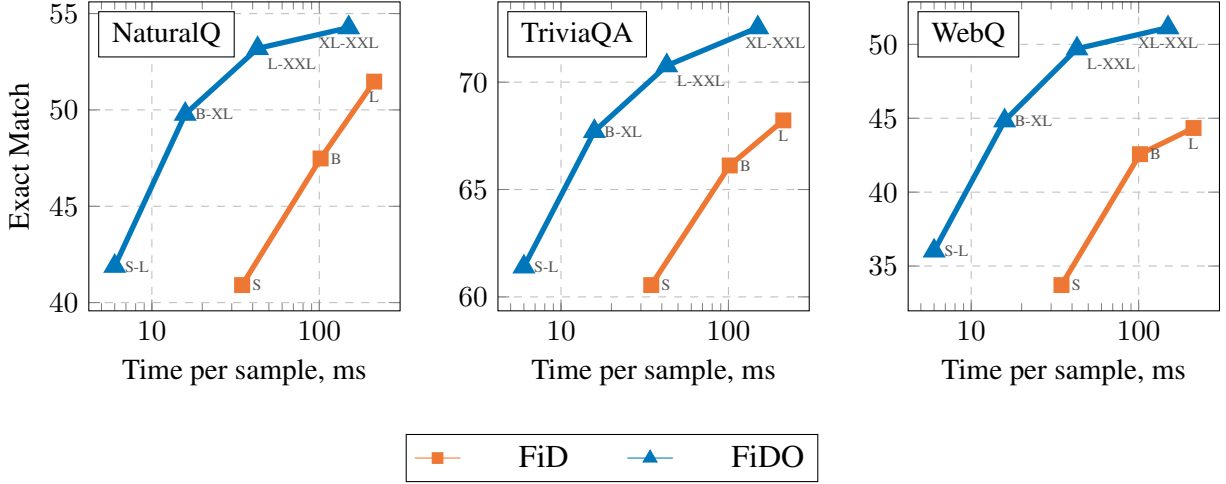
\begin{figure*}
     \centering
     \begin{subfigure}[t]{0.32\textwidth}
        \raggedright
        \begin{tikzpicture}[scale=1.0]
            \begin{axis}[
            scale only axis,
            width=0.8\textwidth,
            height=0.8\textwidth,
            ylabel={Exact Match},
            % ylabel shift=-1pt,
            xlabel={Time per sample, ms},
            mark=x,
            xmode=log,
            % log ticks with fixed point,
            x tick label style={log ticks with fixed point},
            ymajorgrids=true,
            xmajorgrids=true,
            xminorticks=true,
            grid style=dashed,
            % ymin=30,
            % ymax=44,        
            % ticklabel style = {font=\small}
            legend pos={north west},
            ylabel style={align=left, text height=0.2cm},
            ]
            \addlegendimage{empty legend}\addlegendentry{NaturalQ}
            \addplot[color=fidcolor,mark=square*,mark size=2pt,line width=2] table {
                % small
                34.60595312 40.91
                % base
                101.8305 47.48
                % large
            	212.708625 51.47
            };
            \addplot[color=fidocolor,mark=triangle*,mark size=3pt,line width=2] table {
                % small-large
                6.00157812  41.89
                % base-xl
                15.825875 49.78
            	% large-xxl
            	42.82214062 53.19
            	% xl-xxl
            	149.77625 54.27        	
            };
            \node at (axis cs:6.00157812,41.89) [anchor= west, color=darkgrey] {\tiny{S-L}};
            \node at (axis cs:15.825875,49.78) [anchor= west, color=darkgrey] {\tiny{B-XL}};
            \node at (axis cs:42.82214062,53.19) [anchor= north west, color=darkgrey] {\tiny{L-XXL}};
            \node at (axis cs:149.77625,54.27) [anchor= north, color=darkgrey] {\tiny{XL-XXL}};
            
            \node at (axis cs:34.60595312,40.91) [anchor= west, color=darkgrey] {\tiny{S}};
            \node at (axis cs:101.8305,47.48) [anchor= west, color=darkgrey] {\tiny{B}};
            \node at (axis cs:212.708625,51.47) [anchor= north, color=darkgrey] {\tiny{L}};            
            \end{axis}
        \end{tikzpicture}  
     \end{subfigure} \hspace{0.36cm}
     \hfill
     \begin{subfigure}[t]{0.32\textwidth}
        \centering
        % \caption{TriviaQA}        
        % \begin{tikzpicture}[trim axis left,scale=1.0]
        \begin{tikzpicture}[scale=1.0]
            \begin{axis}[
            scale only axis,
            width=0.8\textwidth,
            height=0.8\textwidth,
            % ylabel={Exact Match},
            % ylabel shift=-3pt,            
            xlabel={Time per sample, ms},
            mark=x,
            xmode=log,
            % log ticks with fixed point,
            x tick label style={log ticks with fixed point},
            ymajorgrids=true,
            xmajorgrids=true,
            xminorticks=true,
            grid style=dashed,
            % ymin=30,
            % ymax=44,        
            % ticklabel style = {font=\small}
            legend pos={north west},
            ylabel style={align=left, text height=0.2cm},
            ]
            \addlegendimage{empty legend}\addlegendentry{TriviaQA}
            \addplot[color=fidcolor,mark=square*,mark size=2pt, line width=2] table {
                % small
                34.60595312 60.55
                % base
                101.8305 66.12
                % large
            	212.708625 68.22
            };
            \addplot[color=fidocolor,mark=triangle*,mark size=3pt, line width=2] table {
                % small-large
                6.00157812  61.40
                % base-xl
                15.825875 67.71
            	% large-xxl
            	42.82214062 70.77
            	% xl-xxl
            	149.77625 72.54      	
            };
            
            \node at (axis cs:6.00157812,61.40) [anchor= west, color=darkgrey] {\tiny{S-L}};
            \node at (axis cs:15.825875,67.71) [anchor= west, color=darkgrey] {\tiny{B-XL}};
            \node at (axis cs:42.82214062,70.77) [anchor= north west, color=darkgrey] {\tiny{L-XXL}};
            \node at (axis cs:190.77625,72.54) [anchor= north, color=darkgrey] {\tiny{XL-XXL}};
            
            \node at (axis cs:34.60595312,60.55) [anchor= west, color=darkgrey] {\tiny{S}};
            \node at (axis cs:101.8305,66.12) [anchor= west, color=darkgrey] {\tiny{B}};
            \node at (axis cs:212.708625,68.22) [anchor= north, color=darkgrey] {\tiny{L}};            
            \end{axis}
        \end{tikzpicture}  
     \end{subfigure}     
     \hfill
     \begin{subfigure}[t]{0.32\textwidth}
        \raggedleft
        \begin{tikzpicture}[scale=1.0]
            \begin{axis}[
            scale only axis,
            width=0.8\textwidth,
            height=0.8\textwidth,
            % ylabel={Exact Match},
            % ylabel shift=-3pt,
            xlabel={Time per sample, ms},
            mark=x,
            xmode=log,
            % ytick = {46, 50, 54},
            % log ticks with fixed point,
            x tick label style={log ticks with fixed point},
            ymajorgrids=true,
            xmajorgrids=true,
            xminorticks=true,
            grid style=dashed,
            % ymin=30,
            % ymax=44,        
            % ticklabel style = {font=\small}
            legend pos={north west},
            ylabel style={align=left, text height=0.2cm},
            ]
            \addlegendimage{empty legend}\addlegendentry{WebQ}
            \addplot[color=fidcolor,mark=square*,mark size=2pt, line width=2] table {
                % small
                34.60595312 33.71062992
                % base
                101.8305 42.56889764
                % large
            	212.708625 44.34055118
            };
            \addplot[color=fidocolor,mark=triangle*,mark size=3pt, line width=2] table {
                % small-large
                6.00157812  36.02362205
                % base-xl
                15.825875 44.83267717
            	% large-xxl
            	42.82214062 49.70472441
            	% xl-xxl
            	149.77625 51.13188976
            };
            
            \node at (axis cs:6.00157812,36.02362205) [anchor= west, color=darkgrey] {\tiny{S-L}};
            \node at (axis cs:15.825875,44.83267717) [anchor= west, color=darkgrey] {\tiny{B-XL}};
            \node at (axis cs:42.82214062,49.70472441) [anchor= north west, color=darkgrey] {\tiny{L-XXL}};
            \node at (axis cs:149.77625,51.13188976) [anchor= north, color=darkgrey] {\tiny{XL-XXL}};
            
            \node at (axis cs:34.60595312,33.71062992) [anchor= west, color=darkgrey] {\tiny{S}};
            \node at (axis cs:101.8305,42.56889764) [anchor= west, color=darkgrey] {\tiny{B}};
            \node at (axis cs:212.708625,44.34055118) [anchor= north, color=darkgrey] {\tiny{L}};
            \end{axis}
        \end{tikzpicture}  
     \end{subfigure}   
     \hfill
     
     \begin{tikzpicture}
        \begin{customlegend}[
            legend columns=2,
            legend style={
                align=center,
                column sep=4ex,
                font=\large,
            },
            legend entries={FiD, \modelname}
        ]
            \addlegendimage{mark=square*,solid,color=fidcolor}
            \addlegendimage{mark=triangle*,mark size=3pt,solid,color=fidocolor}   
        \end{customlegend}
    \end{tikzpicture}            
     
    \caption{\textbf{MAIN RESULT. \modelname achieves much higher performance for any given inference budget.} Exact match on Natural Questions (NaturalQ), TriviaQA and WebQuestions (WebQ) test sets as a function of inference budget (log scale). Compares FiD Small, Base and Large models with \modelname Small-Large, Base-XL, Large-XXL and XL-XXL  models.}
    \label{fig:main_result_figure}
\end{figure*}
Retrieval-augmented models generally read many context tokens relative to the number of  question or answer tokens, such that processing retrieved text consumes the bulk of FLOPs. However, past work has shown that most inference time for Fusion-in-Decoder (FiD) is spent in the decoder \citep{fidlight}. Our own experiments support this (Figure \ref{fig:flops_train_inference}). This section investigates FiD's computational structure and decoder inference speed, and finds the slower decoder speed to be the result of memory bandwidth constraints, exacerbated by attention over retrieved documents.

\subsection{Fusion-in-Decoder}

The backbone of the Fusion-in-Decoder model \citep{fid} is a T5 encoder-decoder architecture. The model is provided a question or other input, as well as a number of relevant retrieved text passages. The question is prepended to each retrieved passage, and then the encoder is applied to each passage separately. The resulting representations are concatenated. Finally, the decoder cross-attends to the large number of concatenated representations and assimilates the information from the different passages to generate an answer, hence Fusion-in-Decoder.

\subsection{FLOPs of FiD model} Model speed is determined by the number of FLOPs and the speed at which computations are performed, typically measured in floating point operations per second (FLOP/s). Operations in a Transformer can be roughly divided into MLP layers, attention projection layers, and attention operations. For simplicity, we count only multiplication operations.

Let $d$ be the dimension of the model, $n_s$ the total number of tokens across all passages, $n_p$ the number of tokens in a single retrieved passage, $n_t$ the number of tokens in the target, $L$ the number of layers, and assume the MLP dimension is $4d$. The number of FLOPs %\footnote{For simplicity, we count only multiplication operations.} 
used in an encoder layer is approximately 
\begin{equation*}
    \text{FLOPs}_{\text{enc}L} = \underbrace{8 n_s d^2}_{\text{MLP}} + \underbrace{4 n_s d^2}_{\text{QKVO projections}} + \underbrace{2 n_s n_p d}_{\text{Attention}} 
\end{equation*}
\noindent Since the size of each retrieved passage $n_p \ll d$, computation of the attention score is negligible and we can approximate total FLOPs in the encoder as
\begin{equation}
\label{eqn:encflops}
    \text{FLOPs}_{\text{enc}} \approx 12 n_s d^2 \cdot L
\end{equation}
\noindent Decoder layers additionally have cross-attention layers, leading to FLOPs of
\begin{align*}
    &\text{FLOPs}_{\text{dec}L} = \underbrace{8 n_t d^2 + 4 n_t d^2 + 2 n_t^2d}_{\text{MLP and Self-attention}}\\
    &+ \underbrace{2 n_t d^2}_{\text{Cross-attention QO}} + \underbrace{2 n_s d^2}_{\text{Cross-attention KV}} + \underbrace{2 n_t n_s d}_{\text{Cross-attention}}
\end{align*}
\noindent The output length $n_t \ll n_s, d$, so the only non-negligible term for decoder FLOPs originates from the cross-attention key and value projections, which cost the same FLOPs as encoder key and value projections. We see that the decoder consumes roughly $\frac{1}{6}$ the FLOPs of the encoder.
\begin{equation}
\label{eqn:decflops}
    \text{FLOPs}_{\text{dec}} \approx 2 n_s d^2 \cdot L
\end{equation}

Figure~\ref{fig:flops_train_inference} shows that actual measured training time closely mirrors this FLOPs approximation. However, the decoder is much more expensive for inference. We argue below this is because the decoder is \textit{memory bandwidth constrained} during inference, specifically the cross-attention layers. 

\subsection{Effective computational throughput} 

In order to perform computations, accelerators must transmit data between global memory and registers, which can be a limiting factor. The actual FLOP/s achieved can be usefully modeled with the \textit{roofline} model~\citep{roofline,roofline2,rooflineblog} as the lesser of peak FLOP/s the device is capable of and how fast required data can be transferred.
\begin{align*}
    &\text{Actual FLOP/s} = \min(\text{Peak FLOP/s},\\
    &\underbrace{\text{Operational Intensity}}_{\text{Operations per byte}}\cdot
    \underbrace{\text{Peak Memory Bandwidth}}_{\text{bytes per second}})
\end{align*}
The data constraint is given by the product of device memory bandwidth  -- how fast data can be transferred -- and \textit{operational intensity} -- how many operations are performed per unit of data. The latter is determined by an algorithm's degree of \textit{data reuse}, the number of operations that can be performed before new data needs to be fetched.

High operational intensity is necessary for good performance on modern GPU/TPU hardware, for which peak FLOP/s are usually two orders of magnitude times larger than memory bandwidth \citep{tpuv4, a100}. If operational intensity is too low, the accelerator will spend the majority of its time waiting for data to be transferred to registers. Usually, that happens when the model performs minor computations with large tensors repeatedly, for example in normalization layers or during incremental decoding. 

\subsection{Operational intensity of FiD inference}

\citet{multiquery} shows that the speed of incremental Transformer decoding is memory-bandwidth bound due to low operational intensity. Here we follow their analysis and derive the asymptotic \textit{inverse} of operational intensity -- the ratio of memory operations to the compute performed during each incremental decoding step -- for FiD. Let $b$ be the batch size, $h$ the number of attention heads and assume that attention heads have dimension $\frac{d}{h}$. 

\paragraph{Operational intensity of MLP layer.} For each token the linear projections perform $O(bd^2)$ operations, and load $O(bd + d^2)$ memory, where $bd$ corresponds to activations and $d^2$ to the weight matrices. During training, sequence length effectively multiplies batch size as weights need to be loaded only once for the entire sequence, but for inference each token is processed incrementally. The inverse operational intensity is then
\begin{equation}
    \mathcal{R}^{\text{MLP}} = \frac1{b} + \frac1{d}
\end{equation}
Therefore, obtaining high operational intensity of MLP layer ($\mathcal{R}^{\text{MLP}} \ll 1$) during inference requires a large batch size. 

\paragraph{Operational intensity of attention layers.} Memory bandwidth is a more severe bottleneck for attention inference, particularly cross-attention. At each decoding step the model applies projections for a single token, and has to load all cached key and value projections from encoder tokens and prior decoder tokens into memory. This leads to very low operational intensity. 

Specifically, query/key/value/output projections for a single position take $O(bd^2)$ operations. As discussed earlier, we can ignore the attention computation itself. The model needs to load projection matrices ($O(d^2)$ memory) and past keys and values ($O(bnd)$ memory). Therefore, the inverse operational intensities for self-attention layers, $\mathcal{R}^{\text{S-MHA}}$ and cross-attention layers $\mathcal{R}^{\text{C-MHA}}$ are 
\begin{align}
    \mathcal{R}^{\text{S-MHA}} &= \frac1b + \frac{n_t}{d},& 
    \mathcal{R}^{\text{C-MHA}} &= \frac1b + \frac{n_s}{d} 
\end{align}

Because the source input length $n_s$ is extremely long for FiD, the cross-attention operational intensity is very low, which bottlenecks inference. 

\section{Method}

We have shown that the encoder accounts for the bulk of FiD FLOPs and training cost, while FiD spends the majority of inference time in the decoder due to low operational intensity of cross-attention layers. Next we propose several ways to alleviate the decoder bottleneck. This allows us to efficiently allocate more compute to the decoder by scaling decoder size without significantly increasing the inference speed. We denote Fusion-in-Decoder with the proposed optimizations as \modelname (Fusion-in-Decoder Optimized).

\begin{table}[t!]
\centering
\begin{tabular}{lc}
    \textbf{Model} & \textbf{Max Batch Size} \\
    \toprule
     Vanilla FiD & 24 \\
     + LSA & 128 \\
     + MQ & 256 \\
     + XL Decoder & 128 \\
    \bottomrule
\end{tabular}
\caption{Maximum batch size for QA inference with 40 retrieved passages on a single TPUv4 for FiD Base models with different \modelname components.}
\label{table:batch_size}
\end{table}
\begin{table}[ht!]
\centering
\begin{tabular}{lcc}
    \textbf{Model} & \textbf{Pre-training} & \textbf{Finetuning}\\
    \toprule
     Vanilla FiD & 219.9 & 9.7 \\
     + LSA & 247.0 & 11.8 \\
     + MQ & 248.0 & 11.8\\
     + XL Decoder & 81.9 & 6.9\\
    \bottomrule
\end{tabular}
\caption{Pre-training and fine-tuning samples per second per chip for FiD Base model with varying \modelname components. We use 64 TPUv4 chips and batch size 2048 for pre-training and 32 chips and batch size 64 for fine-tuning. See Section \ref{section:experiment_setup} for training information.}
\label{table:training_speed}
\end{table}

\subsection{Layer-sparse cross-attention}

\begin{figure}  
\centering \hspace{-10pt}
\begin{tikzpicture}[scale=1.0]
\begin{axis}[
    xbar stacked,
    % bar shift=0pt,
    enlarge y limits=0.25,
    width=\columnwidth,
    height=0.6\columnwidth,
    major y tick style = transparent,    
    xmajorgrids = true,
    xlabel = {Time per sample, ms},
    symbolic y coords={LSA-12, LSA-6, LSA-3, no LSA},
    ytick = data,
    xmin=0,
    axis y line*=none,
    axis x line*=bottom,
]
    \addplot[style={decodercolor,fill=decodercolor,mark=none}]
        coordinates {(1.34,no LSA) (1.31,LSA-3) (1.36,LSA-6) (1.36,LSA-12)};
    \addplot[style={crosscolor,fill=crosscolor,mark=none}]
        coordinates {(4.02,no LSA) (1.39,LSA-3) (0.67,LSA-6) (0.33,LSA-12)};        
\end{axis}
    
\begin{customlegend}[
    legend columns=1,
    legend cell align=left,
    legend style={
        anchor=north east, % <-- inside the legend style
        align=left,
        at={(6, 1.5)},
        column sep=1ex
    },
    legend entries={Cross-attention, Decoder other}
]
    \addlegendimage{mark=square*,only marks,solid,color=crosscolor}   
    \addlegendimage{mark=square*,only marks,solid,color=decodercolor}    
    
\end{customlegend}
\end{tikzpicture}            

\caption{Cross-attention and total decoder inference time for \modelname Base-XL with varying factors of layer-sparse cross-attention. The main \modelname configuration uses LSA-6 which has cross-attention every 6 layers.}
\label{table:cross_attention_prop}
\end{figure}
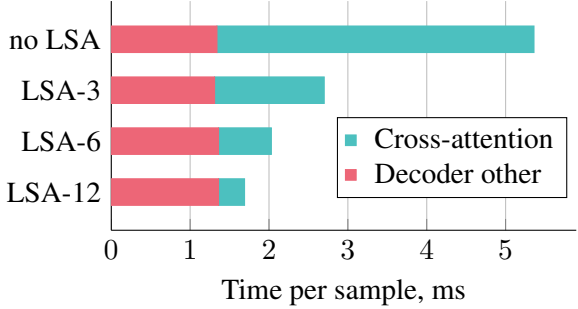

The decoder cross-attention layer is the primary bottleneck for inference due to its low operational intensity. FiD-Light \citep{fidlight} improves the operational intensity by reducing the effective input length by a factor of $K$. We instead propose to remove cross-attention from some decoder layers entirely, keeping cross-attention only in one out of every $K$ decoder layers. We call this layer-sparse cross-attention (LSA). Section \ref{section:experiments} provides evidence that LSA achieves similar speedups without FiD-Light's drop in quality. For \modelname we use LSA with sparsity $K=6$, which means that a Large decoder has cross-attention only at layers 6, 12, 18 and 24. In principle LSA and FiD-Light can be combined, but we find that after applying LSA and multi-query attention the remaining cross-attention makes up a small proportion of decoder inference cost and further speedups from reducing cross-attention are modest (Figure \ref{table:cross_attention_prop}).

Removing cross-attention layers also reduces FiD's FLOPs and memory usage. Cross-attention layers make up approximately $\frac{1}{7}$ of total FiD FLOPs (see Eqn \ref{eqn:decflops}) and applying LSA-6 leads to a 12\% reduction in FLOPs. Table \ref{table:training_speed} shows the reduction in FLOPs is reflected by an increase in training speed. Moreover, cross-attention keys and values make up a substantial proportion of memory usage during inference, and LSA-6 enables a much larger batch size (Table \ref{table:batch_size}).

\subsection{Multi-query attention}

\citet{multiquery} proposes to increase the operational intensity of decoder attention layers by applying multi-query attention, in which keys and values share a single head each and only queries have multiple heads. With a single head, keys and values use a factor $h$ less memory and are much faster to load. With multi-query attention, keys and values occupy $O(bnd/h)$ memory, so that the inverse operational intensity of cross-attention becomes
\begin{equation}
    \mathcal{R}^{\text{C-MQA}} = \frac1b + \frac1d + \frac{n_s}{dh}
    \label{eqn:mqopint}
\end{equation}
which has the problematic term $\frac{n_s}{d}$ reduced by factor of $h$. Multi-query attention further reduces inference cost (Figure~\ref{fig:encoder_decoder_inference}) and memory (Table \ref{table:batch_size}) on top of layer-sparse cross-attention, though not training speed (Table \ref{table:training_speed}).

\subsection{Asymmetric Decoder}

Section \ref{sec:analysis} showed that the FiD encoder consumes an order of magnitude more FLOPs than the decoder because the encoder and decoder are the same size but the encoder is applied to many more tokens. After applying layer-sparse cross-attention and multi-query attention, the decoder also takes up much less time for inference. Such an allocation may not be optimal, as the FiD decoder is responsible for a more challenging task than the standard T5 encoder: it has to assimilate and reason over information from many passages.

We propose to partially redress this imbalance through massively scaling the decoder up, by as much as 15x. Because the decoder is applied to fewer tokens, and because increased decoder dimension improves operational efficiency, such scaling only modestly increases inference cost. For example, Figure~\ref{fig:encoder_decoder_inference} shows that replacing the Base-sized decoder with an XL-sized decoder increases the total inference time per sample by only 21\%. Fine-tuning costs also increase only modestly (Table \ref{table:training_speed}). However, pre-training costs increase more (though still much less than the scaling factor of the decoder), as T5 pre-training uses a much smaller ratio of input length to output length. After reducing the decoder cross-attention memory costs scaling the decoder only mildly increases activation memory, so that \modelname can still fit much larger batch sizes than vanilla FiD (Table \ref{table:batch_size}). For the \modelname method we use decoders that are typically two T5 sizes larger than the encoder: Small-Large, Base-XL, Large-XXL and XL-XXL (as XXL is the largest T5 model). 

\begin{table*}[ht!]
\centering
\begin{tabular}{lcc|ccc}
    \textbf{Model} & \textbf{Total TPS} & \textbf{Decoder TPS} & \textbf{NaturalQ} & \textbf{TriviaQA} & \textbf{WebQ} \\
    \toprule
    \modelname (base-XL) & 15.8  & 2.0 & 48.2 & 67.3 & 46.8 \\
    no LSA & 19.2 & 5.4 & 47.9  & 67.4 & 46.3  \\
    no MQ & 60.8  & 47.0 & 48.2  & 67.5 & 45.4 \\
    no Asym (base-base) & 14.4 & 0.6 & 46.3 & 64.9 & 41.0  \\    
    \bottomrule
\end{tabular}
\caption{Inference time per sample, decoder time per sample (ms) and downstream QA exact match for \modelname base-XL with different components ablated separately. \modelname is evaluated on dev sets for ablation results.}
\label{table:ablations}
\end{table*}

\section{Related Work}

\paragraph{Retrieval-augmented models} There exists a large body of retrieval-augmented approaches. Some particularly well known models are REALM \citep{realm}, RAG \citep{rag}, RETRO \citep{retro} and Fusion-in-Decoder \citep{fid}. FiD in particular has achieved state-of-the-art performance on a wide variety of tasks \citep{fid, atlas, generateratherretrieve} and in this work we focus on improving the performance-efficiency trade-offs for FiD. RETRO is another closely related retrieval-augmented model, as it uses a small encoder for retrieved context and a larger primary decoder like \modelname does. Unlike RETRO, \modelname's efficiency improvements allow it to tractably attend to many retrieved passages with a much larger decoder.

\paragraph{Efficient Transformers} Our work builds heavily on existing insights into neural network and particularly Transformer speed. Previous work has found that data movement is often a constraining factor for computations on modern devices \citep{roofline, flashattention, multiquery}. \citet{multiquery} shows that autoregressive Transformers are particularly bandwidth bound during inference, and proposes multi-query attention as a partial solution. We find that this is exacerbated by the FiD setting, and adopt multi-query attention for \modelname to ameliorate the problem. \citet{palminference} also investigates multi-query attention, primarily in the context of efficient inference and parallelization for very large language models, whereas we focus on performance/cost trade-offs for the retrieval-augmented setting.

Another way to alleviate memory bandwidth constraints is to quantize model parameters and possibly activations \citep{int8, glm}. Quantizing models reduces data that needs to be sent to device registers, and also reduces overall memory usage which allows for larger, more efficient batch sizes. Finally, it is possible to distill \citep{distill, distillsurvey} models into a smaller student model, which is cheaper for inference. However, knowledge distillation requires labeling a very large number of samples with the larger model, so reducing the inference costs of larger models is highly valuable. 

\paragraph{Efficient retrieval-augmented models} \modelname lies in a body of work that attempts to improve the efficiency of retrieval-augmented or long-input models. One direction focuses on reducing the cost of the attention mechanism. LongT5 \citep{longt5} routes long-range attention through a small number of global tokens. FiD-Light \citep{fidlight}, the most closely related work to \modelname,  employs a similar mechanism for FiD, as the decoder attends to only the first $\frac{1}{K}$ proportion of representations of each retrieved passage. We opt to introduce sparsity in attention layers as in ReadTwice \citep{readtwice} instead of attention patterns. \modelname applies cross-attention from the decoder to the encoder in one out of every K layers, which achieves a similar speedup to FiD-Light but with only minor performance penalty. \modelname also incorporates multi-query attention leading to a further order of magnitude reduction in decoder inference cost, and takes advantage of this to massively scale the decoder.

A different and complementary direction is to reduce the cost of reading retrieved passages. KG-FiD \citep{kgfid} reranks retrieved passages and reads only the top passages, while \citet{canext} reads more retrieved passages only if it is not confident in its answer. Another approach is to pre-compute and store encoder representations in a memory and directly retrieve representations from memory, rather than re-encoding retrieved text \citep{tome, memorizing, fidmemory}. For standard FiD, the decoder actually makes up the bulk of the inference cost. \modelname reduces the cost of the decoder such that encoding retrieved passages becomes the bottleneck, increasing the benefit of the above approaches.

\section{Experiments}
\label{section:experiments}

\subsection{Experiment Setup}
\label{section:experiment_setup}

\paragraph{Pre-training} All models are based on the T5.1.1 architecture \citep{t5}, pre-trained from scratch on C4 \citep{c4} using JAX \citep{jax}, FLAX \citep{flax}, and T5X \citep{t5x}. We employ the standard T5 training recipe except for a modified Adafactor \citep{adafactor} optimizer. Appendix \ref{apppendix:training} describes training in greater detail. 

\paragraph{Downstream evaluation}

We evaluate \modelname on open-domain question-answering datasets Natural Questions \citep{nq}, TriviaQA \citep{triviaqa} and WebQuestions \citep{webquestions}. We report results on the open-domain QA splits from \citet{orqa}. For all datasets, each sample is paired with a set of 100-word Wikipedia passages ranked by DPR \citep{dpr} score. The question is prepended to each retrieved passage, and then truncated to 256 tokens. The experiments in the paper use 40 retrieved passages to balance performance and speed, but our results hold across a wide range of retrieved passages.

\paragraph{Inference setup}

For our main results we choose a setting that we believe is most representative for common use of retrieval-augmented models. We perform inference on a single TPUv4 and report inference time per sample (TPS) as measured by xprof \citep{xprof}. We use a batch size of 64 (or the largest batch size that fits, if smaller) for the main experiments. Figure \ref{fig:flops_train_inference} and \ref{fig:encoder_decoder_inference} use batch size 24 to ensure a like-for-like comparison, as it is the largest batch size that fits for vanilla FiD. All experiments use 40 passages of 256 tokens and output size of 32 tokens. Predictions are generated with greedy decoding as we found beam search did not meaningfully improve performance for considered tasks. Analysis in Section \ref{sec:analysis} investigates how trade-offs change with input and output length, low batch size and different sampling methods.

\subsection{Main results}

Figure \ref{fig:main_result_figure} shows performance as a function of inference time for FiD and \modelname. \modelname strongly outperforms FiD at any inference budget and achieves the same performance with order of magnitude faster speed. The following section investigates how each component of \modelname contributes to its performance. Table \ref{table:test_results} compares \modelname to published results.

\subsection{Components}

\paragraph{Layer-sparse cross-attention}
\begin{table}[ht!]
\centering
\begin{tabular}{lc|ccc}
    \textbf{Model} & \textbf{TPS} & \textbf{NQ} & \textbf{TQA} & \textbf{WebQ} \\
    \toprule
    FiD & 101.8 & 46.5 & 65.8 & 41.83 \\
    FiD-Light & 28.3 & 36.3 & 54.5 & 30.8 \\
    FiD-LSA & 29.5 & 45.8 & 65.3 & 41.0 \\
    \bottomrule
\end{tabular}
\caption{Time per sample (ms) and QA exact match for FiD, FiD-Light, and FiD Base-sized models with layer-sparse cross-attention.}
\label{table:layersparse_vs_light}
\end{table}

First, Table \ref{table:ablations} shows that layer-sparse cross-attention significantly reduces inference cost with modest performance degradation. Separately, Table \ref{table:layersparse_vs_light} compares the inference speed and performance impact of layer-sparse cross-attention with the token-sparse cross-attention from FiD-Light. Reducing cross-attention layers and inducing encoder output sparsity by the same factor lead to similar speedups, but layer-sparse cross-attention achieves the inference speedup with much lower performance penalty. 

Note that we find a much larger performance degradation from compressing the encoder output in our setting compared to the experiments in \citet{fidlight}. Some exploratory experiments suggest that multi-task training fine-tuning on large amounts of data as done in FiD-Light may ameliorate the performance penalty from compressing encoder output; however even with such training \citet{fidlight} still report significant peformance degradation, in contrast to LSA.

Layer-sparsity over a factor of 6 incurs greater performance penalties. However, as shown in Table \ref{table:cross_attention_prop}, with LSA-6 cross-attention already makes up a small proportion of total decoder inference cost.

\paragraph{Multi-query attention}

Table \ref{table:ablations} shows that multi-query attention achieves a large cost reduction on top of layer-sparse cross-attention with minimal performance degradation, consistent with our analysis and findings from \citet{multiquery}.

\paragraph{Decoder scale}
\begin{figure}
     \centering
        \begin{tikzpicture}[scale=1.0]
            \begin{axis}[
            scale only axis,
            width=0.8\columnwidth,
            ylabel={Exact Match},
            xlabel={Time per sample (ms, log scale)},
            mark=x,
            xmode=log,
            ymajorgrids=true,
            xmajorgrids=true,
            xminorticks=true,
            grid style=dashed,
            legend columns=1,
            legend cell align=left,
            legend pos={north west},
        ]
            \addplot[color=fidocolor,mark=triangle*,mark size=3pt, line width=1] table {
                % small-large
                6.00157812  42.37
                % base-xl
                15.825875 48.17
            	% large-xxl
            	42.82214062 51.22
            	% xl-xxl
            % 	149.77625 60.2   	
            };
            \addplot[color=red,mark=*,mark size=2pt, line width=1] table {
                5.28 40.2
                % base
                14.38 46.51
                % large
            	37.67 49.51
            	% xl

            };
            \node at (axis cs:6.00157812,42.37) [anchor= south, color=darkgrey] {\small{Small-Large}};            
            \node at (axis cs:15.825875,48.17) [anchor= east, color=darkgrey] {\small{Base-XL}};
            \node at (axis cs:42.82214062,51.22) [anchor= east, color=darkgrey] {\small{Large-XXL}};            

            \node at (axis cs:5.28,40.2) [anchor= west, color=darkgrey] {\small{Small}};            
            \node at (axis cs:14.38,46.51) [anchor= north west, color=darkgrey] {\small{Base}};            
            \node at (axis cs:37.67,49.51) [anchor= north, color=darkgrey] {\small{Large}};            
            
             \legend{\modelname, FiD + LSA + MQ}        
            \end{axis}
        \end{tikzpicture}
    \caption{Performance on Natural Questions dev set as a function of inference time for \modelname Small, Base and Large models with and without asymmetric decoder.}
    \label{fig:asymmetric}
\end{figure}
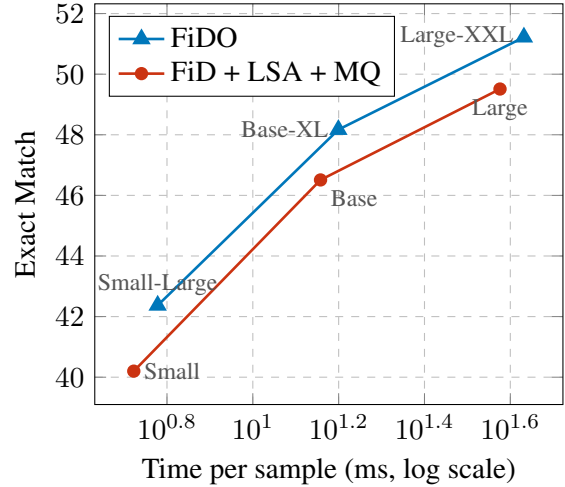
\begin{table}[ht!]
\centering
\small
\begin{tabular}{l|ccc}
    \textbf{Model} & \textbf{NQ} & \textbf{TQA} & \textbf{WQ} \\
    \toprule

    REALM \citep{realm} & 40.4 & - & 40.7 \\    
    RAG \citep{rag} & 44.5 & 56.8 & 45.2 \\
    RETRO \citep{retro} & 45.5 & - & - \\
    T5-XXL \citep{t5ssm} & 35.2 & 51.9 & 42.8 \\
    ATLAS \citep{atlas}  & 60.4 & 79.8 & - \\    
    \midrule
    FiD-L \citep{fid}  & 51.4 & 67.6 & - \\
    FiD-L (ours) & 51.5 & 68.2 & 44.3 \\
    \modelname (L-XXL)  & 53.2 & 70.7 & 49.7 \\
    \bottomrule
\end{tabular}
\caption{Comparison of \modelname with published  results on Natural Questions, TriviaQA and WebQuestions test sets. We focus on comparing with FiD as other works enhance performance with improved retrieval (such as ATLAS), which is orthogonal to our contributions.}
\label{table:test_results}
\end{table}

We can see in Table \ref{table:ablations} that increasing the size of the decoder leads to a significant improvement in performance at the cost of a modest increase in inference time. Figure \ref{fig:asymmetric} provides a visual comparison of the performance-inference profile for \modelname with and without asymmetric decoders and shows that asymmetric large decoders achieve a better trade-off.

\subsection{Other analysis}
\label{sec:analysis}
\paragraph{Varying input and target length}
\begin{figure}
    \centering
    \begin{subfigure}[t]{0.23\textwidth}
        \centering
        \caption{TPS by \# of retrievals}
        \hspace{-0.36cm}
        \begin{tikzpicture}[scale=1.0]
            \begin{axis}[
            draw,
            scale only axis,
            width=0.87\columnwidth,
            xlabel style = {font=\small},
            mark=x,
            ymode=log,
            xtick={20, 60, 100},
            % x tick label style={log ticks with fixed point},
            y tick label style={/pgf/number format/sci},
            % log ticks with fixed point,
            ymajorgrids=true,
            xmajorgrids=true,
            xminorticks=true,
            grid style=dashed,
            label style={text height=6.0pt},
            ticklabel style = {font=\small,text width=10pt}
            ]
            \addplot[color=fidcolor,mark=square*,mark size=2pt] table {
                10 25
                20 50
                40 102
                60 156
                80 203
                100 256
            };
            \addplot[color=cyan,mark=diamond*,mark size=3pt] table {
                10 8
                20 15
                40 30
                60 45
                80 58
                100 73
            };
            \addplot[color=red,mark=*,mark size=2pt] table {
                10 4 
                20 7
                40 14
                60 21
                80 29
                100 36
            };
            \addplot[color=fidocolor,mark=triangle*,mark size=3pt] table {
                10 5 
                20 9
                40 16
                60 23
                80 30
                100 38
            };
            \end{axis}
            % \draw [red] (current bounding box.south west) rectangle (current bounding box.north east);
        \end{tikzpicture}
     \end{subfigure} \hspace{-0.15cm}
     \begin{subfigure}[t]{0.23\textwidth}
        \centering
        \caption{TPS by output length}
        \begin{tikzpicture}[scale=1.0]
            \begin{axis}[
            width=0.87\columnwidth,
            scale only axis,
            xlabel style = {font=\small},
            mark=x,
            ymode=log,
            ytick={100, 1000},
            xtick={32, 256, 512},
            % xmode=log,
            % log ticks with fixed point,
            % y ticks = {100, 1000},
            ymajorgrids=true,
            xmajorgrids=true,
            xminorticks=true,
            grid style=dashed,
            ticklabel style = {font=\small,text width=10pt},
            label style={text height=4.0pt},
            ]
            \addplot[color=fidcolor,mark=square*,mark size=2pt] table {
                32 102
                64 187
                128 357
                256 684
                512 1404
            };
            \addplot[color=cyan,mark=diamond*,mark size=3pt] table {
                32 30
                64 45
                128 75
                256 139
                512 279
            };
            \addplot[color=red,mark=*,mark size=2pt] table {
                32 14
                64 15
                128 16
                256 18
                512 24
            };
            \addplot[color=fidocolor,mark=triangle*,mark size=3pt] table {
                32 16
                64 18
                128 21
                256 33
                512 56
            };
            \end{axis}
            % \draw [red] (current bounding box.south west) rectangle (current bounding box.north east);
        \end{tikzpicture}
    \end{subfigure}
    \hfill
    \begin{tikzpicture}
        \begin{customlegend}[
            legend columns=2,
            legend style={
                align=center,
                column sep=1ex,
            },
            legend entries={FiD, FiD + LSA, FiD + LSA + MQ, \modelname}
        ]
            \addlegendimage{mark=square*,solid,color=fidcolor}
            \addlegendimage{mark=diamond*,solid,color=cyan}
            \addlegendimage{mark=*,solid,color=red}
            \addlegendimage{mark=triangle*,mark size=3pt,solid,color=fidocolor}   
        \end{customlegend}
    \end{tikzpicture}            

    \caption{Time per sample (TPS) as a function of retrieved passages (left) or the number of generated tokens (right) for Base FiD variants and \modelname-Base-XL.}
    \label{fig:speed_vs_length}
\end{figure}
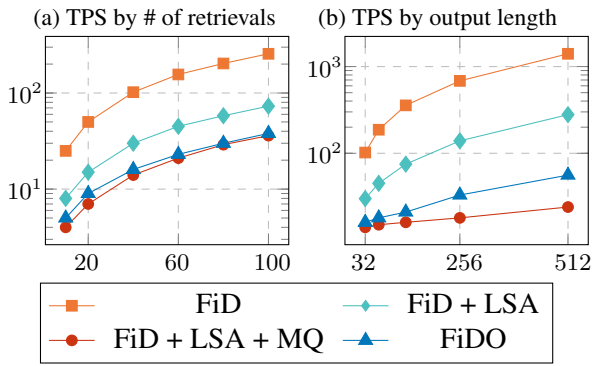

Our main results use a middle-of-the-road setting for FiD applications with a medium number of retrievals and a relatively short output, reflecting common knowledge-intensive tasks. However, it is interesting to ask how \modelname components affect speed for other settings. Figure \ref{fig:speed_vs_length} shows time per sample as a function of retrieved passages and length of the target output for each step from FiD to \modelname. 

We first note that layer-sparse cross-attention and multi-query attention are critical across all settings. For standard output length, the asymmetric decoder is cheap for any reasonable number of retrieved passages, becoming negligible as a fraction of total inference time as the number of retrievals increases. As output length increases, the cost of the disproportionately large decoder rises, although it only becomes a substantial proportion of inference time for output length of 256-512 and above. For tasks with long outputs, such as summarization, one may want to reduce the level of decoder asymmetry (e.g. Base-Large rather than Base-XL).

\paragraph{Low batch size setting}

For our primary investigation we focus on medium batch sizes (24+). There are two reasons one might care about smaller batch sizes: either because larger batches do not fit in memory or because they lead to excessive latency. The first constraint is not binding for \modelname: due to \modelname's memory efficiency we are able to fit larger batches even for the XL-XXL model, and if necessary model size can be further extended with quantization \citep{glm} and parallelism \citep{palminference}. 

For real-time serving latency can be a constraint, but in those settings it is common practice to use much smaller models which are distilled from larger teacher models \citep{distillsurvey}. The student models can utilize a higher batch size, while the teacher models do not have latency constraints, so \modelname also applies to this use case.

For rare cases where a lower batch size is required layer-sparse and multi-query attention are still important, but cannot fully eliminate the decoder as a bottleneck for inference (Table \ref{table:low_batch_size}). The $\frac{1}{b}$ term in Equation \ref{eqn:mqopint} dominates, reflecting the fact that the model has to repeatedly load model parameters without spreading the cost over many samples.

Instead of scaling the decoder, it would be more cost-effective to apply more expensive sampling methods, because sampling methods increase the effective batch size. For example, beam search with large beams is nearly free at lower batch sizes.

\begin{table}[ht!]
\centering
\begin{tabular}{lcc}
    \textbf{Model} & \textbf{Total TPS} & \textbf{Decoder TPS} \\
    \toprule
     Vanilla FiD &  135  & 123 \\
     + LSA & 51 & 39  \\
     + MQ & 35 & 23 \\
     + Beam 16 & 35 & 23 \\
     + XL Decoder & 117 & 105 \\
    \bottomrule
\end{tabular}
\caption{Inference time per sample (ms) with batch size 1 for Base FiD with varying \modelname components.}
\label{table:low_batch_size}
\end{table}

\paragraph{Sampling}

We do not apply beam search for our main experiments as decoder inference time is proportional to beam width for medium batch sizes and beam search does not improve performance on the considered set of tasks. Instead, we find that scaling decoder size provides a more cost-efficient way to add decoder capacity. Table \ref{table:beam} compares the performance vs time trade-offs from beam search and scaling the decoder for Natural Questions, and shows that scaling the decoder is significantly more effective. Beam search may be more important for other tasks, such as tasks with longer outputs.

\begin{table}[ht!]
\centering
\begin{tabular}{lcc}
    \textbf{Model} & \textbf{Decoder TPS} & \textbf{NaturalQ} \\
    \toprule
     FiD with LSA, MQ &  0.6  & 46.3 \\
     + Beam 4 & 2.4 & 46.2 \\
     \modelname & 2.0 & 48.2 \\
    \bottomrule
\end{tabular}
\caption{Decoder inference time (ms) and QA exact match for FiD Base models, comparing the trade-offs of beam search versus scaling decoder size.}
\label{table:beam}
\end{table}

\section{Conclusion}

We perform analysis of the performance-inference speed tradeoff for FiD, showing that the encoder uses more FLOPs but most time is spent in the decoder due to memory bandwidth constraints. We propose \modelname, an extension of FiD which removes most cross-attention layers and employs multi-query attention to vastly reduce the cost of the decoder. The resulting model spends most time in the encoder, consistent with compute analysis, which \modelname takes advantage of by strongly increasing the size of the decoder. We show that \modelname achieves much stronger performance for the same inference budget relative to existing FiD models.

\section*{Acknowlegements}
We thank Livio Baldini Soares, Kenton Lee, Pat Verga, Iftekhar Naim and others at Google Research for insightful advice and discussion. Michiel de Jong is partially supported by NSF Awards IIS-1513966/ 1632803/1833137, CCF-1139148, DARPA Awards\#: FA8750-18-2-0117, FA8750-19-1-0504,  DARPA-D3M - Award UCB-00009528, Google Research Awards, gifts from Facebook and Netflix, and ARO\# W911NF-12-1-0241 and W911NF-15-1-0484.

\section*{Limitations}

One of the advantages of the Fusion-in-Decoder approach is that it uses the off-the-shelf T5 architecture with publicly available checkpoints. The proposed \modelname modifications strongly improve performance and inference speed for retrieval-augmented question-answering, but require pre-training from scratch. It is in general preferable to have a small number of checkpoints that can be fine-tuned for any application. For example, it may not be feasible to train different giant language models for use in the retrieval-augmented setting. Instead, the architectures for such large models may need to be a compromise for different use cases.

\section*{Ethics}

In general the ethics concerns for this paper are similar to those for the large body of work studying retrieval-augmented language models. One distinction worth pointing out is that this work proposes a model with faster inference, which makes retrieval-augmented models more feasible to apply in practical settings and serve to users and inherently carries higher risk.

\bibliographystyle{acl_natbib}
\bibliography{anthology,custom}

\appendix

\section{Training}
\label{apppendix:training}
All experiments are built on the T5.1.1 architecture with the training recipe from T5 \citep{t5}. The first exception is the optimizer; we find that the second moment factoring and mixing schedule from Adafactor \citep{adafactor} can lead to instability, especially with unbalanced encoder and decoder sizes. Instead, we disable factoring and second moment mixing, leading to an optimizer that is a hybrid between Adafactor and Adam \citep{adam}.

The second difference to the training recipe arises from the observation that \modelname XL-XXL is unstable for the standard training regimen. We solve the instability by restarting from a recent healthy checkpoint with a 10x decreased learning rate, which happened once. 

During fine-tuning, we load not only model weights but also second moment estimates, which we find leads to better fine-tuning in general and particularly for asymmetric models. We finetune with learning rate 0.001 and batch size 64 for all datasets. For evaluation on test sets we select the checkpoint with the best validation performance.
\newpage

\end{document}